\documentclass[letterpaper, 10 pt, conference]{ieeeconf}  

\IEEEoverridecommandlockouts                              

\usepackage{amssymb}            
\usepackage{bbold}
\usepackage{mathtools}          
\usepackage{mathrsfs}           
\usepackage{graphicx}           
\usepackage[space]{grffile}     
\usepackage{overpic}
\usepackage{hyperref}
\usepackage{url}                
\usepackage{lipsum}             
\usepackage{multirow}
\usepackage{xcolor}

\DeclareMathOperator*{\argmax}{arg\,max}

\title{\LARGE \bf
Learning to Drive by Imitating Surrounding Vehicles
}

\author{Yasin Sonmez, Hanna Krasowski and Murat Arcak
\thanks{Authors are with Faculty of Electrical Engineering and Computer Sciences,
        University of California, Berkeley, CA, USA
        {\tt\small yasin\_sonmez@berkeley.edu}}
}

\begin{document}

\maketitle
\thispagestyle{empty}
\pagestyle{empty}

\begin{abstract}
    Imitation learning is a promising approach for training autonomous vehicles (AV) to navigate complex traffic environments by mimicking expert driver behaviors. 
    While existing imitation learning frameworks focus on leveraging expert demonstrations, they often overlook the potential of additional complex driving data from surrounding traffic participants. In this paper, we study a data augmentation strategy that leverages the observed trajectories of nearby vehicles, captured by the AV’s sensors, as additional demonstrations. We introduce a simple vehicle-selection sampling and filtering strategy that prioritizes informative and diverse driving behaviors, contributing to a richer dataset for training. We evaluate this idea with a representative learning-based planner on a large real-world dataset and demonstrate improved performance in complex driving scenarios. Specifically, the approach reduces collision rates and improves safety metrics compared to the baseline. Notably, even when using only 10 percent of the original dataset, the method matches or exceeds the performance of the full dataset. Through ablations, we analyze selection criteria and show that naive random selection can degrade performance. Our findings highlight the value of leveraging diverse real-world trajectory data in imitation learning and provide insights into data augmentation strategies for autonomous driving.

\end{abstract}

\section{INTRODUCTION}
By learning from expert demonstrations, imitation learning enables autonomous vehicles (AVs) to develop policies that mimic human-like driving behavior. Recently, imitation learning models \cite{pluto, zheng2025diffusion} have started to outperform traditional rule-based methods \cite{dauner2023parting} on benchmarks with large-scale real-world data such as nuPlan \cite{caesar2021nuplan}, indicating the increasing viability of imitation learning for real-world deployment.
However, imitation learning also suffers from three major challenges. First, recent studies demonstrate that imitation learning models can learn shortcuts from data \cite{jaeger2023hidden}, leading to undesired behaviors. For example, it has been demonstrated that models with historical AV motion data excel in open-loop evaluation but underperform in closed-loop metrics, likely due to learning shortcuts \cite{cheng2024rethinking}. Second, imitation learning suffers from the distribution shift problem, where the training and test sets have different distributions due to the nature of the application, such as learning from data collected in one location and deploying the model elsewhere. To address this challenge, several studies suggest that imitation learning benefits from reinforcement learning refinements \cite{lu2023imitation}. Lastly, imitation learning suffers from causal confusion \cite{de2019causal} when a model learns spurious correlations instead of true causal relationships between actions and outcomes. Since imitation learning relies on mimicking expert demonstrations, the model may pick up on irrelevant features or unintended cues that correlate with successful behavior but do not actually cause it. 

Addressing these challenges through effective data augmentation, model architecture, and loss choices is essential for improving real-world performance. As such, we have to maximize the utility of the available data. Despite the availability of large datasets, simulators, and benchmarks (e.g., \cite{caesar2021nuplan, gulino2023waymax, dauner2025navsim}), effectively utilizing this data for imitation learning remains a challenge. Different datasets capture driving information at varying levels of abstraction, ranging from object-level annotations to raw sensor images. Furthermore, recent studies have indicated that simply increasing the volume of training data does not necessarily result in improved model performance. For example, \cite{pmlr-v205-bronstein23a} highlights that more data may not always translate to better outcomes, suggesting that other factors, such as data quality and relevance, play a more significant role in model effectiveness.

In this paper, we study a data augmentation technique that enhances imitation learning for autonomous driving by using trajectories beyond those of the AV in driving datasets. Building on the idea of learning from other vehicles in a scene, we analyze this strategy in the object-based planning setting. We introduce simple selection criteria that prioritize informative and diverse driving trajectories from observed vehicles and validate the approach through extensive ablation studies. The main contributions of this work are:
\begin{itemize}
    \item We demonstrate the effectiveness of learning from surrounding vehicles on a large real-world object-based planning dataset (nuPlan), unlike prior studies that are limited to simulation environments. Our approach leverages real human driving behaviors from surrounding vehicles rather than simulated trajectories, providing more realistic and diverse training signals.
    \item We conduct extensive ablations on vehicle-selection criteria and analyze how different selection strategies affect learning performance, addressing a key limitation of prior research that assumes all vehicles are useful for learning.
\end{itemize}

\begin{figure*}[tb]
    \centering
    \begin{overpic}[width=\linewidth]{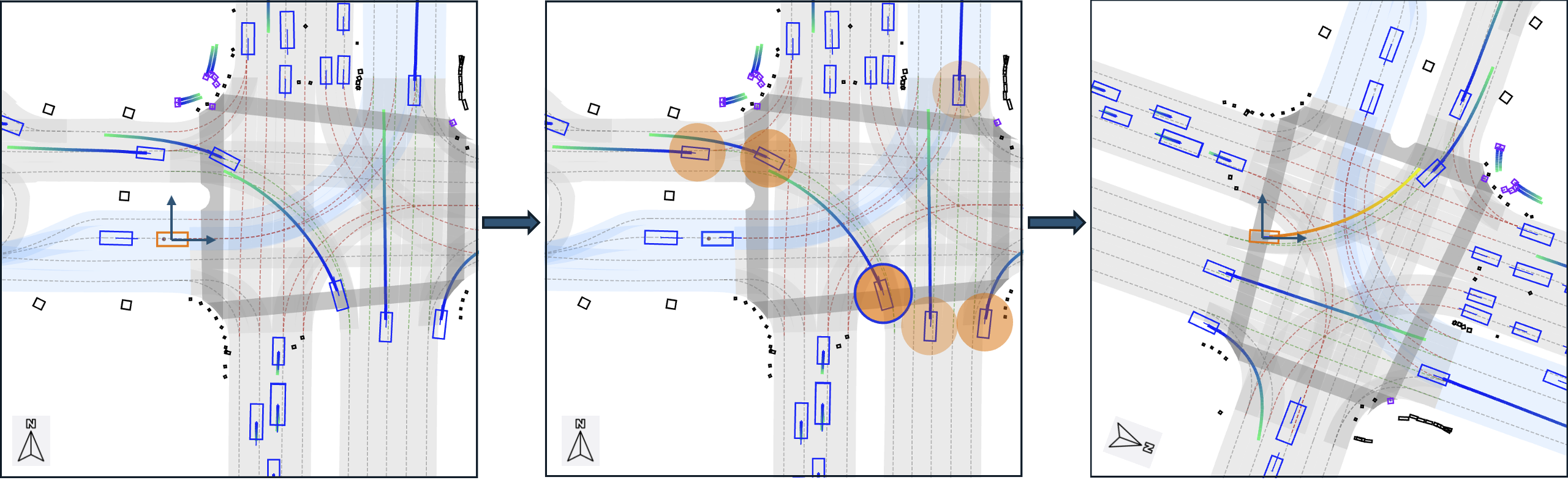}
        \put(5,-4){\textbf{1. Original Data}}
        \put (7, -10){$\mathbf{x}=\begin{bmatrix} F_{AV} \\ F_o \\ \vdots \end{bmatrix}$}

        \put(41,-4){\textbf{2. Vehicle Selection}}
        \put(41,-12){$i^* \sim \mathtt{Softmax}(I, \mathbf{h})$ }

        \put(75,-4){\textbf{3. Augmented Data}}
        \put (77, -10){$\mathbf{\tilde{x}}=\begin{bmatrix} \tilde{F}_{AV} \\\tilde{F}_o \\ \vdots\end{bmatrix}$}
    \end{overpic}
        \vspace{21mm}
    \caption{Data augmentation framework illustrated for a traffic scenario: (1) The expert driver remains stationary at a red light while surrounding vehicles follow diverse trajectories. (2) A suitable surrounding vehicle $i^*$ (shown with a blue circle) is sampled from the weighted categorical distribution defined in Equation \ref{eq:sampling}, where darker circles represent higher selection probabilities and $\mathbf{h}$ is the vector of sum of absolute heading deviations for all vehicles $I$. (3) A reference frame transformation generates features from the perspective of the new ego vehicle.}
    \label{fig:framework}
\end{figure*}

\section{Related Work}\label{sec:citations}

\textbf{Datasets for autonomous driving:}
Although traffic data is tedious and costly to collect, there is an increasing amount of open-source datasets for autonomous driving research. These datasets can be broadly categorized into perception (i.e., sensor-based \cite{Geiger2013IJRR-KITTI,Argoverse,nuscenes2019}) and motion planning (i.e., object-based \cite{highDdataset, interactiondataset, caesar2021nuplan, BARMPOUNAKIS202050}) datasets. Since imitation learning is commonly used for motion planning tasks, we focus on object-based datasets, which have the additional advantage of allowing for interpretable and complete data augmentation. The object-based dataset NGSIM \cite{NGSIM} is one of the earlier large datasets initially published in 2006. Since then many new datasets with increasing size and traffic complexity have been published. In particular, the nuPlan benchmark \cite{karnchanachari2024towards, caesar2021nuplan} consists of real-world autonomous driving datasets and evaluation frameworks. nuPlan offers a comprehensive dataset for both prediction and planning, with 1282 hours of driving data from four cities, and introduces a taxonomy of driving scenarios. Due to these features, nuPlan has been used to compare various planning approaches in the literature such as \cite{huang2023gameformer, dauner2023parting, sharan2023llm}. 

\textbf{Data augmentation for autonomous driving: } Although above datasets are of increasing size, data augmentation can significantly enhance their value. For example, \cite{Guo_2024_CVPR} develops context-aware data augmentation for imitation learning that is based on a variational autoencoder. PLUTO \cite{pluto} employs contrastive imitation learning to address distribution shift by applying both positive and negative data augmentations, where positive augmentations agree with the ground truth and negative augmentations intentionally disagree. 

Recent studies have also started exploring leveraging trajectories of surrounding vehicles for data augmentation. Chen et al.~\cite{Chen2022LearningFA} propose learning from all vehicles observed by the ego-vehicle in a simulator, using simulated agents with simple trajectories to increase sample efficiency. Zhang et al.~\cite{zhang2021learning} present a "Learning by Watching" approach that converts observations into Bird's-Eye-View representations and infers actions from observed vehicles. However, these approaches suffer from several limitations: (1) they are restricted to simulation environments where surrounding vehicles exhibit simplified, rule-based behaviors that may not capture real human driving complexity; (2) they lack intelligent vehicle selection mechanisms, using all observed vehicles without considering that many surrounding vehicles may exhibit uninformative or harmful behaviors; (3) they lack evaluation on large-scale real-world datasets, making it difficult to assess practical effectiveness.

\textbf{Imitation learning for autonomous driving}:
The two main learning approaches for autonomous driving are reinforcement learning and imitation learning. Reinforcement learning usually relies on a realistic simulation environment and significant reward-shaping to achieve performant driving policies \cite{Kiran2021-survey}. Imitation learning is usually easier to tune but requires a diverse and large dataset of driving trajectories to achieve expert-like driving behavior \cite{LeMero2022-survey, Ly2021-survey}. Leading companies in autonomous driving, such as Tesla and Waymo \cite{bansal2018chauffeurnet, lu2023imitation}, as well as open-source projects like OpenPilot \cite{openpilot}, leverage imitation learning to train models by mimicking expert driving behavior. A recent example is the work by \cite{zheng2025diffusion}, who propose a transformer-based Diffusion Planner for closed-loop planning, capable of modeling multi-modal driving behavior. Another notable framework in imitation learning is PLUTO \cite{pluto}, which introduces key innovations for more efficient driving behavior generation: a longitudinal-lateral aware transformer architecture, contrastive learning to mitigate causal confusion and distribution shift, and ego-related data augmentation. In this study, we use PLUTO as a baseline and perform ablation studies on the nuPlan dataset.

\section{Methodology}
\label{sec:methodology}
\subsection{Problem Formulation}

Our method is a general data augmentation approach applicable to object-based planning frameworks. Below, we briefly formulate a generic object-based planning problem that the planner used in our experiments also follows.

We consider an AV, \( N_A \) dynamic agents, \( N_O \) static obstacles, a high-definition map \( M \), and traffic context information \( C \) (e.g., traffic light status). Each agent \( i \) at time \( t \) has a state \( \mathbf{s}_i^t = (\mathbf{p}_i^t, \theta_i^t, \mathbf{v}_i^t, \mathbf{b}_i^t, I_i^t) \), where \( \mathbf{p}_i^t \in \mathbb{R}^2 \) and \( \theta_i^t \in \mathbb{R} \) denote position and heading, \( \mathbf{v}_i^t \in \mathbb{R}^2\) represents velocity, and \( \mathbf{b}_i^t \in \mathbb{R}^2\) and \( I_i^t\in \{0,1\} \) correspond to the bounding box dimensions and observation status, respectively. The feature set for dynamic agents is denoted as \( \mathcal{A} = A_{0:N_A} \), where \( A_0 \) represents the AV, and the static obstacle set is \( \mathcal{O} = O_{1:N_O} \). The future state of agent \( a \) at time \( t \) is denoted as \( \mathbf{y}_a^t \), with historical and future horizons of \( T_H \) and \( T_F \), respectively. The planner generates \( N_T \) multi-modal planning trajectories for the AV along with predictions for each dynamic agent. The final trajectory \( \tau^* \) is selected via a scoring module \( S \), which integrates these outputs with the scene context. The overall formulation is given as:
\begin{equation}
(\mathbf{T}_0, \mathbf{\pi}_0), \mathbf{P}_{1:N_A} = f(A, \mathcal{O}, M, C \mid \phi)
\end{equation}
\begin{equation}
    (\tau^*, \pi^*) = \argmax_{(\tau, \pi) \in (\mathbf{T_0}, \mathbf{\pi_0})} S(\tau, \pi, \mathbf{P}_{1:N_A}, \mathcal{O}, M, C),    
\end{equation}
where \( f \) represents the planning model, \( \phi \) are the model parameters, \( (\mathbf{T}_0, \mathbf{\pi}_0) = \{(\mathbf{y}_{0, i}^{1:T_F}, \pi_i) \mid i = 1, \dots, N_T\} \) are the generated planning trajectories with confidence scores, and \( \mathbf{P}_{1:N_A} = \{\mathbf{y}_a^{1:T_F} \mid a = 1, \dots, N_A\} \) are the predicted future states of traffic participants.

\subsection{Learning From Surrounding Traffic}

Many real-world driving scenarios are inherently dynamic, with multiple interacting agents. While some situations may involve routine behaviors such as lane-keeping or waiting at a red light, even in these types of scenarios usually there is at least one vehicle exhibiting complex or ``interesting" behaviors in the surrounding traffic. Examples include lane changes, turning at intersections, yielding to pedestrians, or reacting to bicycles. These nuanced interactions provide a rich source of data for understanding diverse driving behaviors and the decision-making processes of road users. Learning from surrounding traffic method capitalizes on this by augmenting the imitation learning dataset with estimated trajectories of selected agents from the surrounding traffic. 

The main advantage is increased data diversity along two dimensions. First, incorporating trajectories from surrounding vehicles expands the support of the training distribution by adding non-routine maneuvers (turns, lane changes, yielding), varied driver styles and vehicle dynamics, and exposure to multi-agent interactions. This reduces overfitting, improves generalization \cite{wang2020improving, yu-etal-2022-data}, and improves anticipation in interaction-heavy scenes \cite{zhang2024multi}. Second, prioritizing dynamic and contextually rich cases rather than routine AV-only behavior has been shown to improve performance \cite{pmlr-v205-bronstein23a}. 

\begin{figure}[t!]
    \centering
    \includegraphics[width=0.9\linewidth]{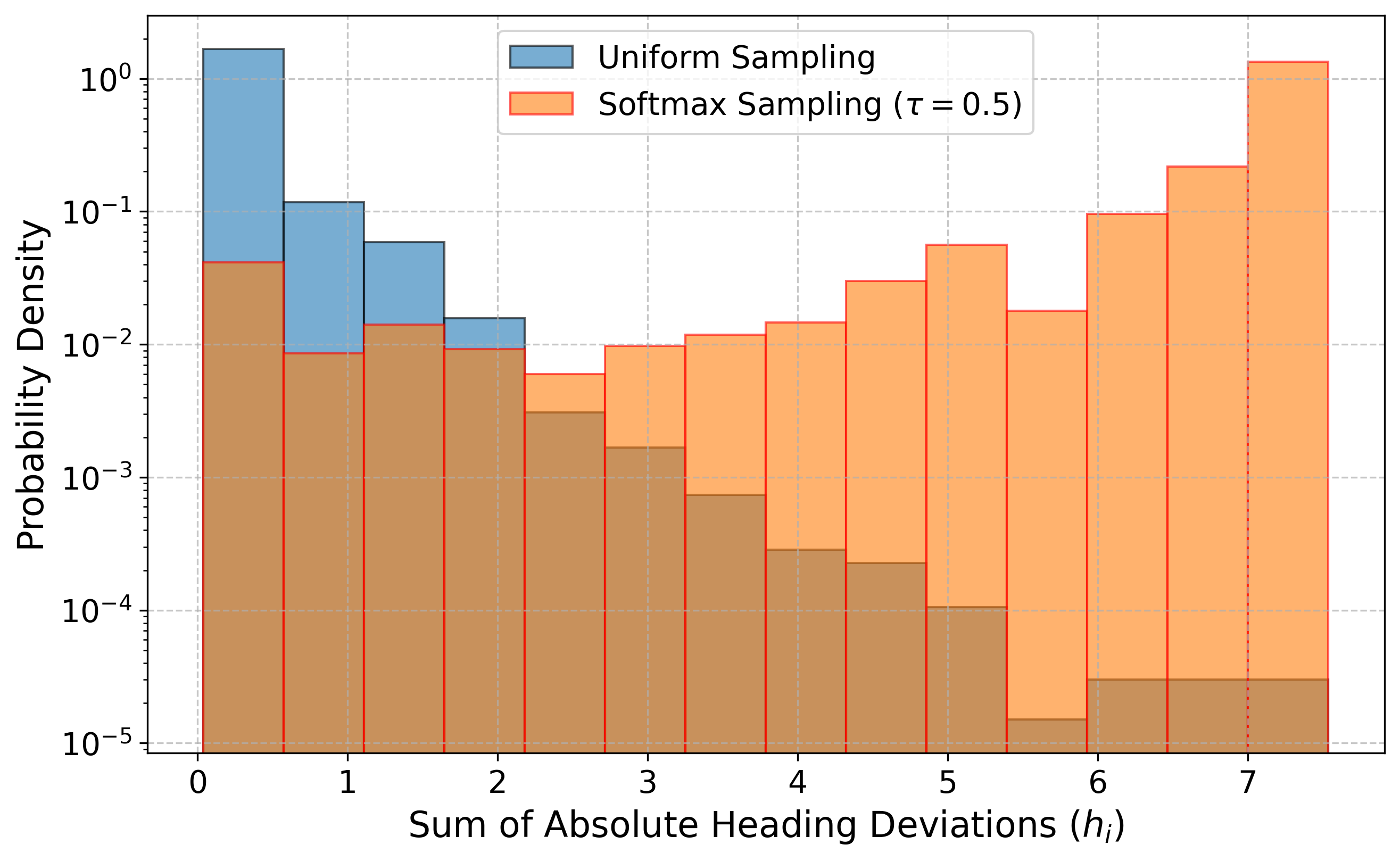}
    \caption{Histogram of the sum of absolute heading deviations $h_i$ for observed vehicles in the dataset. The blue plot represents the original data, while the orange plot corresponds to vehicles sampled using the softmax distribution defined in Equation \ref{eq:sampling}. The histogram is presented on a logarithmic scale to account for the large differences in probabilities.}
    \label{fig:histogram}
\end{figure}

The distribution of the sum of absolute heading angle deviations in nuPlan dataset, illustrated in Figure \ref{fig:histogram}, further supports the need for increased data diversity. The blue histogram indicates that the majority of observed vehicle data consists of minimal deviation movements, such as lane-keeping, simple acceleration, and deceleration. This demonstrates that routine behaviors dominate the dataset, highlighting the importance of our approach to incorporate diverse motion patterns from dynamic surrounding vehicles. 

\subsection{Vehicle Selection Criteria} Most simulation scenarios have many vehicles present, but it is infeasible to augment the dataset with all of them. Many of these vehicles may exhibit little to no interesting behavior, and including them would unnecessarily increase the computation and contribute little to the learning process. Prior research has shown that treating all training data equally can lead to suboptimal performance, especially in safety-critical situations, and that prioritizing more informative samples can improve robustness while reducing data requirements. For instance, \cite{pmlr-v205-bronstein23a} demonstrated that an imitation-learning-based planner trained on only $10\%$ of a dataset, carefully curated using a trained scenario difficulty predictor, performed as well as one trained on the full dataset while significantly reducing collisions and improving route adherence. Therefore, vehicle selection criteria are essential to ensure the dataset contains the most beneficial scenarios. 

Let $s$ index a scene with agent set $\mathcal{I}_s=\{0,1,\dots,N_A\}$ (with $0$ for the AV), time indices $t\in\{0,1,\dots,T_H-1\}$, map $M$, and context $C$. We first define the per-scene pool of eligible non-AV agents via filtering
\begin{equation}
\hat{\mathcal{V}}_s
= \Bigg\{ i\in \mathcal{I}_s \setminus \{0\} :
\begin{aligned}
& I_i^t = 1 \quad && \forall\, t, \\
& \|\mathbf{p}_i^t - \mathbf{p}_0^t\|_2 \le r \quad && \forall\, t, \\
& \mathbf{p}_i^t \in \mathtt{Drivable}(M) \quad && \forall\, t
\end{aligned}
\Bigg\},
\label{filter1}
\end{equation}
where $I_i^t\in\{0,1\}$ indicates observability of the agent in that timestep, $r=50\,\mathrm{m}$, and $\mathtt{Drivable}(M)$ denotes the drivable set induced by the map. This filtering ensures data quality and consistency by: (i) requiring full observability across all timesteps to avoid partial sensing artifacts; (ii) constraining proximity to the AV for reliable measurements; and (iii) ensuring drivability to focus on meaningful driving interactions rather than parked vehicles.


\paragraph{Weighted sampling based on heading deviations} 
Our vehicle selection strategy is based on the observation that vehicles performing complex maneuvers such as turns, lane changes, and parking operations exhibit significant changes in their heading direction, while routine behaviors like lane-keeping and straight-line driving result in minimal heading variations. As shown in Figure~\ref{fig:histogram}, the distribution of heading deviations reveals that most vehicles in the dataset exhibit low deviation values, corresponding to routine driving behaviors that dominate the training data. By prioritizing vehicles with higher heading deviations, we effectively balance the dataset toward more informative and diverse driving patterns that are crucial for learning robust driving policies. While more sophisticated selection criteria based on jerk, acceleration patterns, or driver expertise could be explored, we deliberately use heading deviation as a simple, interpretable, and reasonably effective proxy for dynamic behavior that captures the characteristics of complex driving interactions.

After filtering according to Equation \ref{filter1}, for the remaining pool of vehicles, we assign a weight $h_i$ based on their sum of absolute heading angle deviation over time:
\begin{equation}
    h_i = \sum_t \big|\theta_i^t - \theta_i^{t-1}\big|, \qquad i\in \hat{\mathcal{V}}_s.
\end{equation}
We then define a softmax probability distribution over eligible agents $\hat{\mathcal{V}}_s$ to sample more informative vehicles:
\begin{equation}
    p_{\tau}(i\mid s) 
    = \frac{\exp(h_i/\tau)}{\sum_{j\in \hat{\mathcal{V}}_s} \exp(h_j/\tau)} ,
\label{eq:sampling}
\end{equation}
where $h_i$ represents the weight for vehicle $i$, $p_{\tau}(i\mid s) = \mathtt{Softmax}(i\mid s)$ is its assigned probability, and $\tau$ is the temperature parameter that controls the sharpness of the weight distribution. A lower $\tau$ makes the selection more focused on vehicles with higher deviations, while a higher $\tau$ results in a more uniform weighting across all vehicles.

We explore two different sampling approaches for vehicle selection. The first approach, \emph{per-scene sampling}, selects $N_s$ agents from each scene independently using the softmax distribution in Equation~\ref{eq:sampling}:
\begin{equation}
    \mathcal{I}_s = \{i_1,\dots,i_{N_s}\} \sim \text{SWR}\big(p_{\tau}(\cdot\mid s),\; N_s\big),
\end{equation}
where SWR denotes sampling without replacement. This strategy ensures diversity across different traffic scenarios by forcing selection from each scene. The second approach, \emph{per-ego sampling}, selects one agent per ego-vehicle scenario across the entire dataset; for $N_e$ ego scenarios in the dataset, yielding $N_e$ selected agents, rather than sampling scene by scene. While per-ego sampling may select globally highest-scoring agents, it risks over-representing certain scenarios.

\paragraph{Driving agent filters} Many eligible agents provide little learning signal (e.g., near-stationary) or reflect undesired behavior. To bias sampling toward informative and reliable candidates, we compute three per-agent quantities (displacement, comfort violations, and TTC violations) over $t\in\{0,\dots,T_H-1\}$ from the states $\{\mathbf{p}_i^t,\theta_i^t,\mathbf{v}_i^t\}$ (Section~\ref{sec:methodology}) and they are used to prune the candidates before sampling.

\emph{Displacement} encourages agents that actually move and interact with the scene:
\begin{equation}
    d_i = \big\|\, \mathbf{p}_i^{\,T_H-1} - \mathbf{p}_i^{\,0} \,\big\|_2.
\end{equation}
\emph{Comfort violations} filters highly jerky or unstable motion that is often noisy or undesirable to imitate. Using $\mathbf{v}_i^t$ in $i$'s heading-aligned frame, obtain accelerations $a_{x,i}(t),a_{y,i}(t)$, jerks $\dot a_{x,i}(t),\dot a_{y,i}(t)$, and yaw rate/acceleration $\dot\theta_i(t),\ddot\theta_i(t)$. We define the comfort violation count as
\begin{align}
v_i^{\text{comf}}(t) &= \mathbb{1}\{ |a_{x,i}(t)|>\alpha_x \wedge |a_{y,i}(t)|>\alpha_y \\
&\quad \wedge\ |\dot a_{x,i}(t)|>\beta_x \wedge |\dot a_{y,i}(t)|>\beta_y \\
&\quad \wedge\ |\dot\theta_i(t)|>\gamma_1 \wedge |\ddot\theta_i(t)|>\gamma_2 \},\\
V^{\text{comf}}_i &= \sum_{t=0}^{T_H-1} v_i^{\text{comf}}(t),
\end{align}
\emph{TTC violations} detect agents frequently entering imminent-collision geometry. We use an in-lane, longitudinal approximation along agent $i$'s heading. Let $g_0$ denote the front-to-rear gap between agent $i$'s front bumper and agent $j$'s rear bumper. Let $u_i$ and $u_j$ be the agents' longitudinal speeds along $i$'s heading, and define the closing speed $\Delta u = u_i - u_j$. The in-lane TTC and its violation count are
\begin{align}
\text{ttc}_{i,j}(t)&=\begin{cases}\dfrac{g_0}{\max(\epsilon,\,\Delta u)} & g_0>0,\ \Delta u>0,\\ +\infty & \text{otherwise},\end{cases}\\
V^{\text{ttc}}_i &= \sum_{t=0}^{T_H-1} \mathbb{1}\{ \min_{j\ne i} \text{ttc}_{i,j}(t) < \theta_{\text{TTC}} \}.
\end{align}

\noindent Let $\mathcal{F} \subseteq \{\text{disp},\text{comf},\text{ttc}\}$ denote the set of active filters. The filtered pool used for sampling is
\begin{equation}
\tilde{\mathcal{V}}_s(\mathcal{F})=\Bigg\{ i\in \hat{\mathcal{V}}_s :
\begin{aligned}
& (\text{disp}\in\mathcal{F} \Rightarrow d_i \ge d_{\min}) \\
& \wedge\ (\text{comf}\in\mathcal{F} \Rightarrow V^{\text{comf}}_i \le \kappa_{\text{comf}}) \\
& \wedge\ (\text{ttc}\in\mathcal{F} \Rightarrow V^{\text{ttc}}_i \le \kappa_{\text{ttc}})
\end{aligned}
\Bigg\}.
\end{equation}
Thresholds $(d_{\min},\kappa_{\text{comf}},\kappa_{\text{ttc}})$ and the choice of $\mathcal{F}$ are specified in Section~\ref{sec:experiments}. By default, we sample from $\hat{\mathcal{V}}_s$; in filtering ablations we replace it with $\tilde{\mathcal{V}}_s(\mathcal{F})$.

\begin{figure}[tb!]
    \centering
    \includegraphics[width=0.8\linewidth]{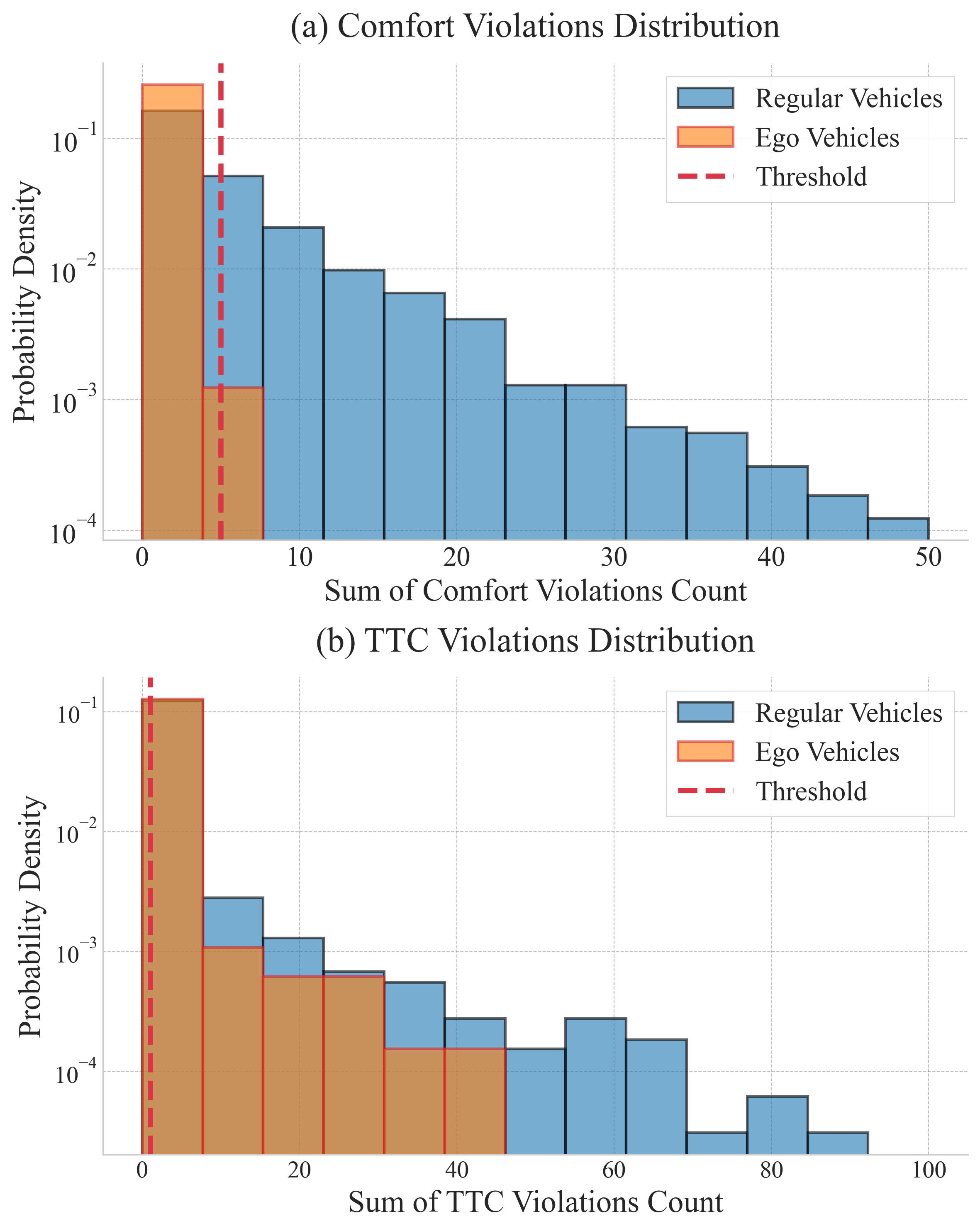}
    \caption{Comparison of TTC and Comfort violation counts for the ego vehicle versus surrounding vehicles across scenes. Ego trajectories tend to have fewer violations, while non-ego trajectories show higher rates due to suboptimal behaviors and greater measurement noise.}
    \label{fig:ego-vs-others}
\end{figure}

To motivate our filtering design, we compare violation rates for the AV versus other traffic participants. Figure~\ref{fig:ego-vs-others} shows that ego trajectories exhibit fewer TTC and comfort violations than surrounding vehicles, either because the ego is a trained expert driver or because non-ego trajectories are noisier due to partial observability and estimation artifacts. This motivates applying simple, robust criteria to prioritize informative and reliable non-ego trajectories.

\subsection{Dataset Augmentation} For each selected agent $i\in \mathcal{I}_s$, let $T_i\in \mathrm{SE}(2)$ denote the rigid transform matrix mapping world coordinates to agent-$i$'s egocentric frame. We apply this transform to all agent spatial components such as position, velocity, and orientation. Denote the original input feature vector by $\mathbf{x}_s$; the augmented feature vector is
\begin{equation}
    \tilde{\mathbf{x}}_{s,i} = f_{T_i}(\mathbf{x}_s),
\end{equation}
where $f_{T_i}$ applies the rigid transform $T_i$ to the spatial components of $\mathbf{x}_s$. Over the dataset $\mathcal{D}=\{\mathbf{x}_s\}_{s=1}^N$, the augmented dataset is
\begin{equation}
    \mathcal{D}_{\text{aug}} 
    = \mathcal{D} \,\cup\, \bigcup_{s=1}^{N} \big\{\tilde{\mathbf{x}}_{s,i} : i\in \mathcal{I}_s\big\}.
\end{equation}

\section{Experiments}

We choose the PLUTO planner \cite{pluto} as the baseline, as it is a high-performance learning-based planner that uses object-based datasets. We selected PLUTO due to its recency, strong closed-loop performance on nuPlan, and compatibility with object-based data. 
While our method can be applied to other similar object-based planning methods such as Diffusion Planner \cite{zheng2025diffusion}, we focused on PLUTO due to computational constraints. Each training and evaluation cycle requires approximately 6 hours for pre-processing, 48 hours for model training on 4 A5000 GPUs, and 6 hours for simulation rollouts. Nevertheless, our data augmentation technique operates independently of PLUTO's internal components (e.g., confidence scores, trajectory scoring, goal specification, model architecture details).
For completeness, the PLUTO planner operates based on:

\textbf{Agent History Encoding: }PLUTO uses the agent state \( \mathbf{s}_i^t \) defined in Section~\ref{sec:methodology}. The trajectories of agents are captured by computing differences between consecutive timesteps, resulting in a feature matrix \( \mathbf{F}_A \in \mathbb{R}^{N_A \times (T_H - 1) \times 8} \).

\textbf{Static Obstacles Encoding: }Static obstacles in the drivable area are encoded as \( \mathbf{o}_i = (\mathbf{p}_i, \theta_i, \mathbf{b}_i) \), producing a feature matrix \( \mathbf{F}_O \in \mathbb{R}^{N_O \times 5} \).

\textbf{AV’s State Encoding: }
To mitigate that, imitation learning frequently learns performance-degrading shortcuts \cite{cheng2024rethinking, wen2020fighting}, only the current state of the AV is used as input features without using the history. These include the AV’s position, heading angle, velocity, acceleration, and steering angle, represented as \( \mathbf{F}_{AV} \in \mathbb{R}^{1 \times 8} \).

\textbf{Vectorized Map Encoding: }The map consists of \( N_p \) polylines, each undergoing an initial subsampling step to standardize the number of points. Feature vectors are then computed for each polyline point. Specifically, for the \( i \)-th point of a polyline, the feature vector consists of $\big(\mathbf{p}_i - \mathbf{p}_0,\, \mathbf{p}_i - \mathbf{p}_{i-1},\, \mathbf{p}_i - \mathbf{p}_{i}^{\text{left}},\, \mathbf{p}_i - \mathbf{p}_{i}^{\text{right}} \big)$
where \( \mathbf{p}_0 \) is the initial point of the polyline, and \( \mathbf{p}_{i}^{\text{left}} \) and \( \mathbf{p}_{i}^{\text{right}} \) represent the left and right lane boundary points, respectively. The final representation of the polyline features is \( \mathbf{F}_P \in \mathbb{R}^{N_P \times n_p \times 8} \), where \( n_p \) is the number of points per polyline.

\textbf{Scene Encoding: } To capture interactions between dynamic agents, static obstacles, polylines, and the autonomous vehicle, all described encodings are concatenated and processed through transformer encoders, with Fourier-based positional embeddings and learnable semantic attributes compensating for the loss of global positional information. 

\textbf{Trajectory Planning and Post-processing: } The model generates multimodal trajectories with confidence scores and employs a rule-based post-processing module to ensure safe and robust selection. Forward simulation, utilizing a linear quadratic regulator for trajectory tracking and a kinematic bicycle model for state updates, assesses rollouts based on metrics such as driving comfort, and time-to-collision. The final trajectory $\tau^*=\arg\max_{(\tau,\pi)\in(\mathbf{T}_0,\mathbf{\pi}_0)} S(\tau,\pi,\mathbf{P}_{1:N_A},\mathcal{O},M,C)$ combines learning-based confidence with rule-based evaluations.

\subsection{Training and Evaluation}
We train the baseline PLUTO planner using varying numbers of scenarios extracted from the nuPlan dataset. For each scenario, we generated $N_s$ additional scenarios from surrounding vehicles. However, in some cases, no suitable vehicles were consistently observable across all time steps, and augmentation was not applied then. All models were trained to convergence, monitored using validation error. We used the hyperparameters of the PLUTO implementation.

For evaluation, we use the test14-hard benchmark, which was curated by executing 100 scenarios for each of 14 scenario types and selecting the 20 lowest-performing instances per type using the rule-based planner PDM-Closed \cite{dauner2023parting}. We do not repeat experiments on val14, a uniformly sampled nuPlan scenarios dataset, due to the strong correlation between learning-based methods' performance on test14-hard and val14, and substantially higher computational costs.

The nuPlan framework provides a comprehensive evaluation score for each simulation, incorporating key metrics such as (1) \textbf{No Ego At-Fault Collisions}, where only AV-initiated collisions are considered; (2) \textbf{TTC (Time-to-Collision) Compliance}, ensuring time-to-collision remains above a threshold; (3) \textbf{Drivable Area Compliance}, requiring the AV to stay within road boundaries; (4) \textbf{Comfort}, assessed via acceleration, jerk, and yaw dynamics within empirical thresholds; (5) \textbf{Progress}, measured as the AV’s traveled distance relative to the expert driver. We use the non-reactive closed-loop score as our performance evaluation metric.\footnote{Thresholds used in our filters follow nuPlan's implementations at \url{https://github.com/motional/nuplan-devkit/blob/master/docs/metrics_description.md}.}

\begin{table*}[bt!]
\caption{Selected performance results across dataset sizes and filtering strategies.}
\label{tab:planner_comparison}
\centering
\small
\renewcommand{\arraystretch}{1.3}
\setlength{\tabcolsep}{5pt}
\begin{tabular}{|c|c|c|c c c c c c|}
\hline
\textbf{Scenarios} & \textbf{Sampling} & \textbf{Filter} & \textbf{Score} & \textbf{Collisions} & \textbf{TTC} & \textbf{Drivable} & \textbf{Comfort} & \textbf{Progress} \\ \hline
\multirow{5}{*}{\textbf{1K}} &  & Baseline & 58.60 & 83.02 & 76.23 & 92.45 & 70.94 & 71.20 \\
 & Per-scene $N_s$ & No filter (N$_s$=1, $\tau$=0.5) & 60.31 & 82.76 & 75.48 & 92.34 & 68.20 & 74.66 \\
 & Per-scene $N_s$ & + Disp & \textbf{65.49} & 84.51 & 74.90 & \textbf{93.73} & \textbf{73.73} & \textbf{76.98} \\
 & Per-scene $N_s$ & + Disp + TTC & 64.17 & \textbf{87.02} & \textbf{78.63} & \textbf{93.51 }& 70.61 & 73.96 \\
 & Per-scene $N_s$ & + Disp + TTC + Comfort & 56.07 & 80.47 & 73.83 & 91.41 & 68.36 & 74.50 \\ \hline
\multirow{5}{*}{\textbf{10K}} &  & Baseline & 61.95 & 83.90 & 74.91 & 93.26 & 76.03 & 78.72 \\
 & Per-scene $N_s$ & No filter (N$_s$=1, $\tau$=0.5) & 72.17 & 91.29 & 82.58 & 95.08 & 80.68 & 80.61 \\
 & Per-scene $N_s$ & + Disp & 73.98 & \textbf{92.80} & 84.44 & 94.55 & \textbf{82.49} & \textbf{81.77} \\
 & Per-scene $N_s$ & + Disp + TTC & \textbf{74.41} & \textbf{92.34} & 84.29 & \textbf{96.55} & 80.84 & \textbf{81.58} \\
 & Per-scene $N_s$ & + Disp + TTC + Comfort & 73.69 & 91.70 & \textbf{86.04}& 94.72 & 81.13 & 80.43 \\ \hline
\multirow{2}{*}{\textbf{100K}} &  & Baseline & 74.81 & 91.23 & \textbf{83.96} & \textbf{97.01} & 86.57 & 78.64 \\
 & Per-scene $N_s$ & No filter (N$_s$=1, $\tau$=0.5) & \textbf{77.38} & \textbf{93.75} & \textbf{84.09} & \textbf{96.97} & \textbf{87.88} & \textbf{80.47} \\ \hline
\end{tabular}\\
\vspace{0.1cm}
\footnotesize{Note: Bold values indicate the highest score in each column for each dataset size.}
\label{tab:planner_comparison}
\end{table*}

\subsection{Results}
We conducted experiments for datasets with 1K, 10K, and 100K scenarios. As shown in Table \ref{tab:planner_comparison}, our data augmentation method consistently outperforms the baseline across all dataset sizes, achieving performance improvements of $11.7\%$, $20.1\%$, and $3.4\%$, respectively. A deeper analysis of individual metrics reveals the most significant enhancements in collision and time-to-collision, indicating that our method substantially reduces the likelihood of collisions and enhances safety. Notably, even when using only $10\%$ of the original dataset, our augmented approach achieves better performance than using the full dataset, outperforming the baseline in terms of both collision rate and TTC. This highlights the effectiveness of our data augmentation strategy in improving both safety and model performance.

\begin{figure}[t!]
    \vspace{0.5cm}
    \centering
    \begin{overpic}[width=0.22\linewidth, trim=300 300 300 300, clip]{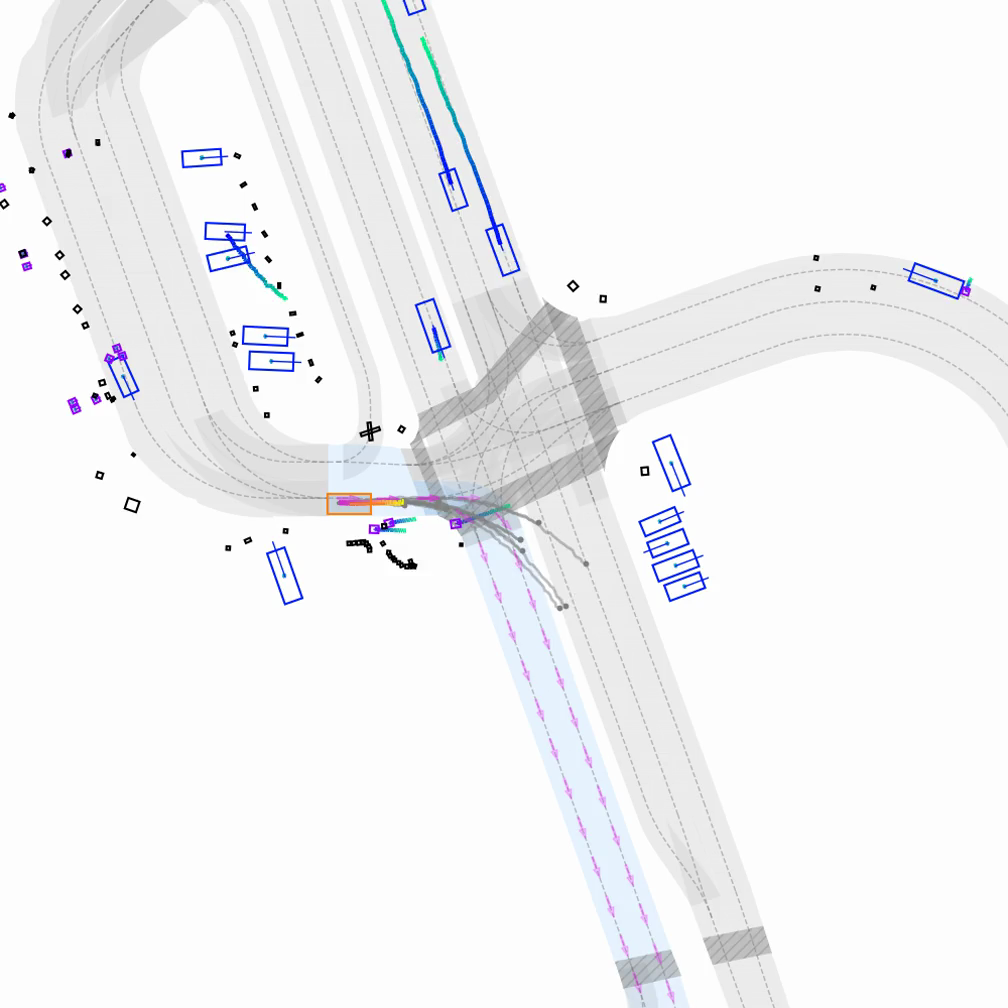}
        \put(25, 105){\textbf{$t = 0s$}} 
        \put(-22, 20){\rotatebox{90}{\text{Baseline}}} 
    \end{overpic}
    \begin{overpic}[width=0.22\linewidth, trim=300 300 300 300, clip]{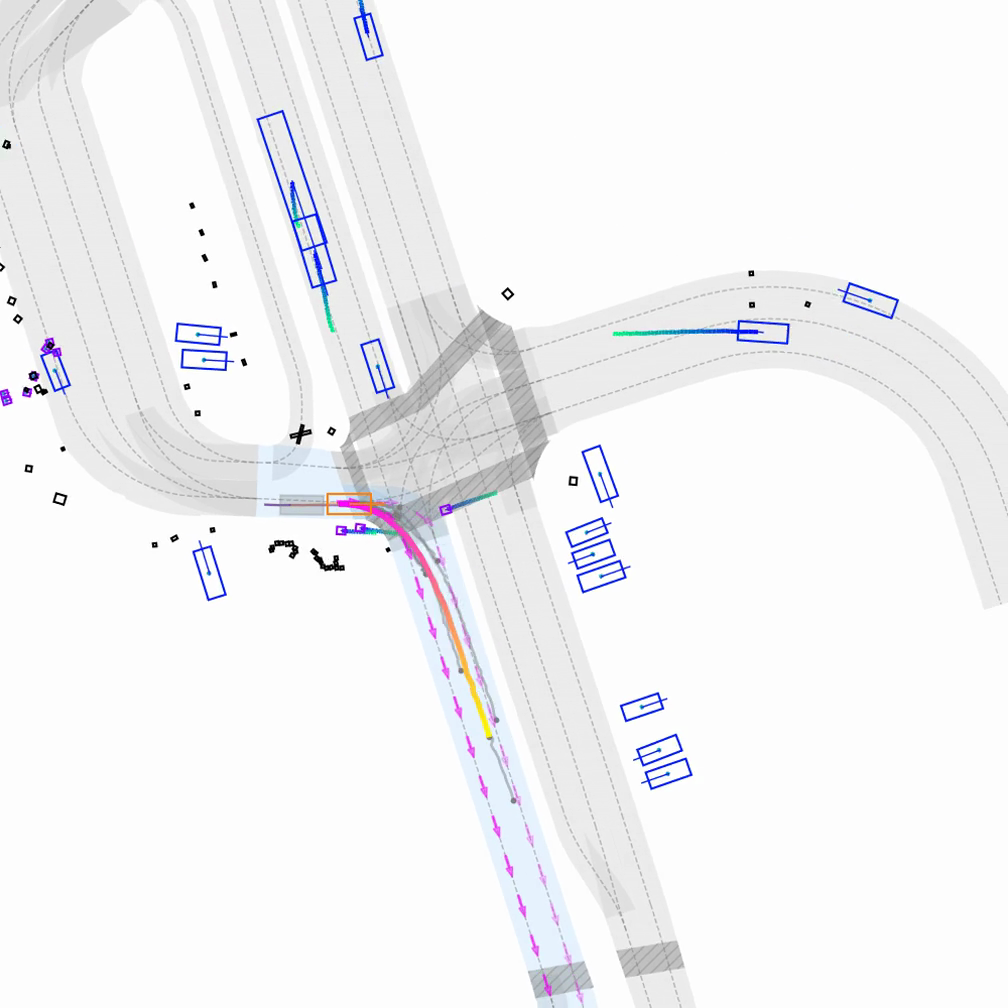}
        \put(25, 105){\textbf{$t = 4s$}} 
    \end{overpic}
    \begin{overpic}[width=0.22\linewidth, trim=300 300 300 300, clip]{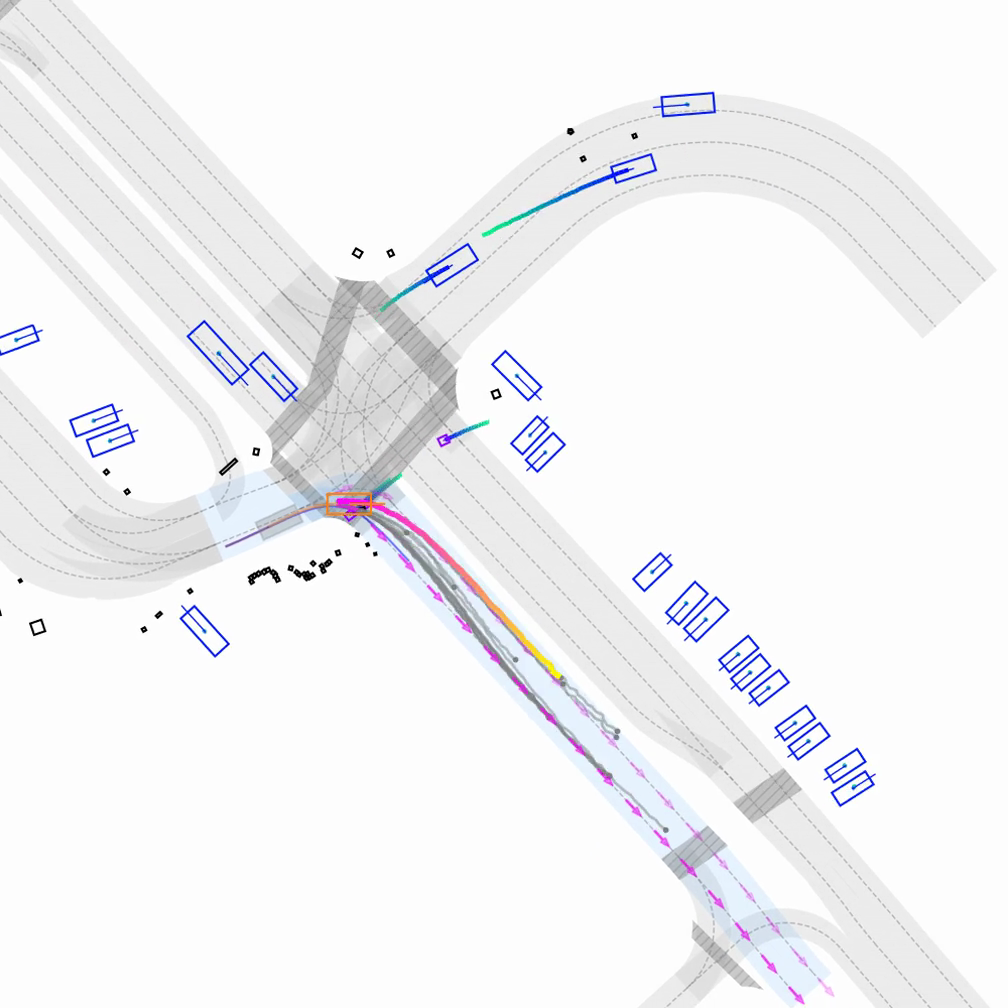}
        \put(25, 105){\textbf{$t = 8s$}} 
    \end{overpic}
    \begin{overpic}[width=0.22\linewidth, trim=200 200 200 200, clip]{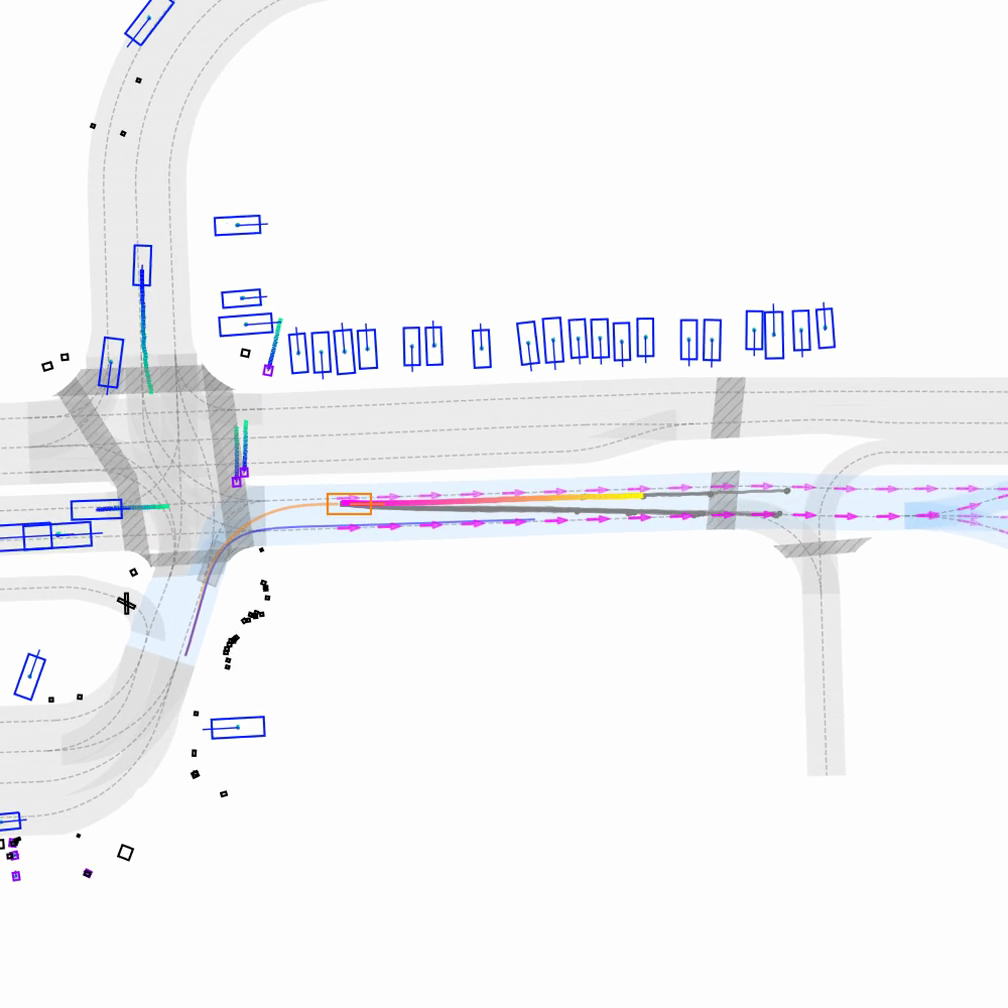}
        \put(25, 105){\textbf{$t = 12s$}} 
    \end{overpic}

    \begin{overpic}[width=0.22\linewidth, trim=300 300 300 300, clip]{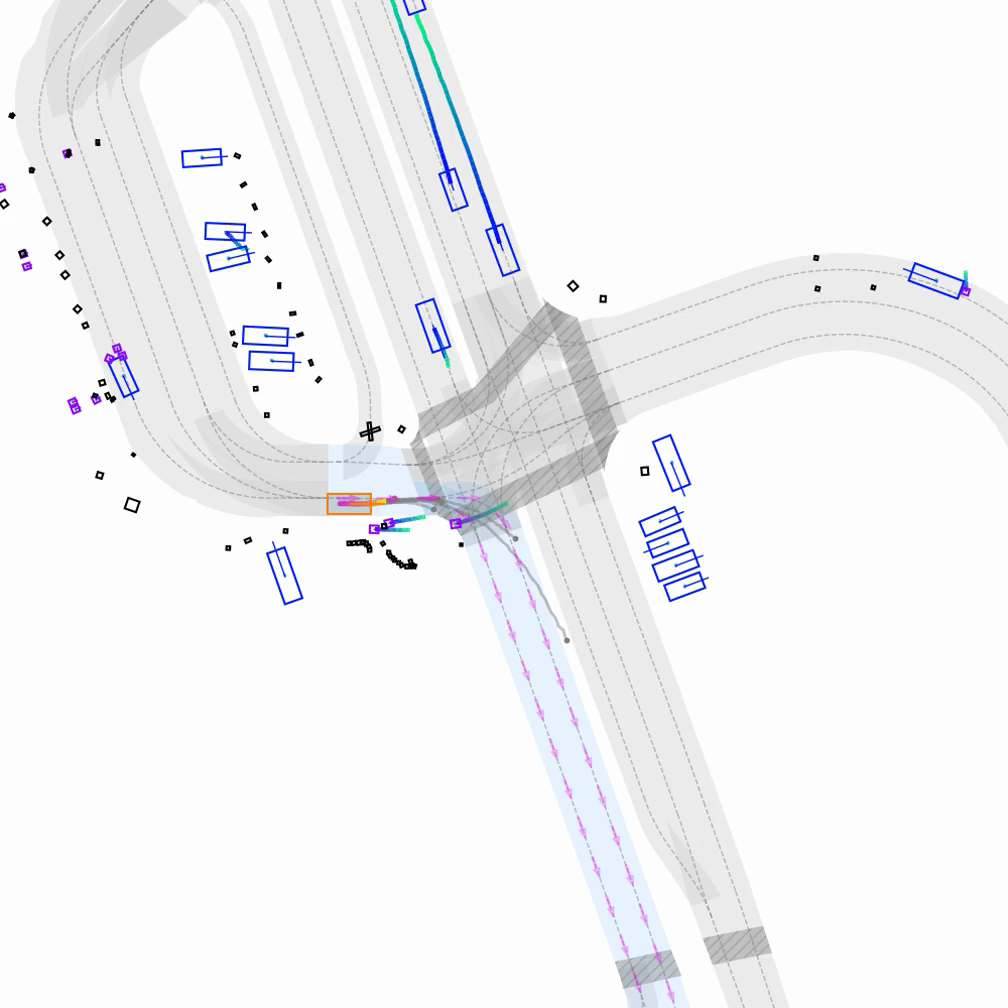}
        \put(-22, 33){\rotatebox{90}{\text{Ours}}} 
    \end{overpic}
    \includegraphics[width=0.22\linewidth, trim=300 300 300 300, clip]{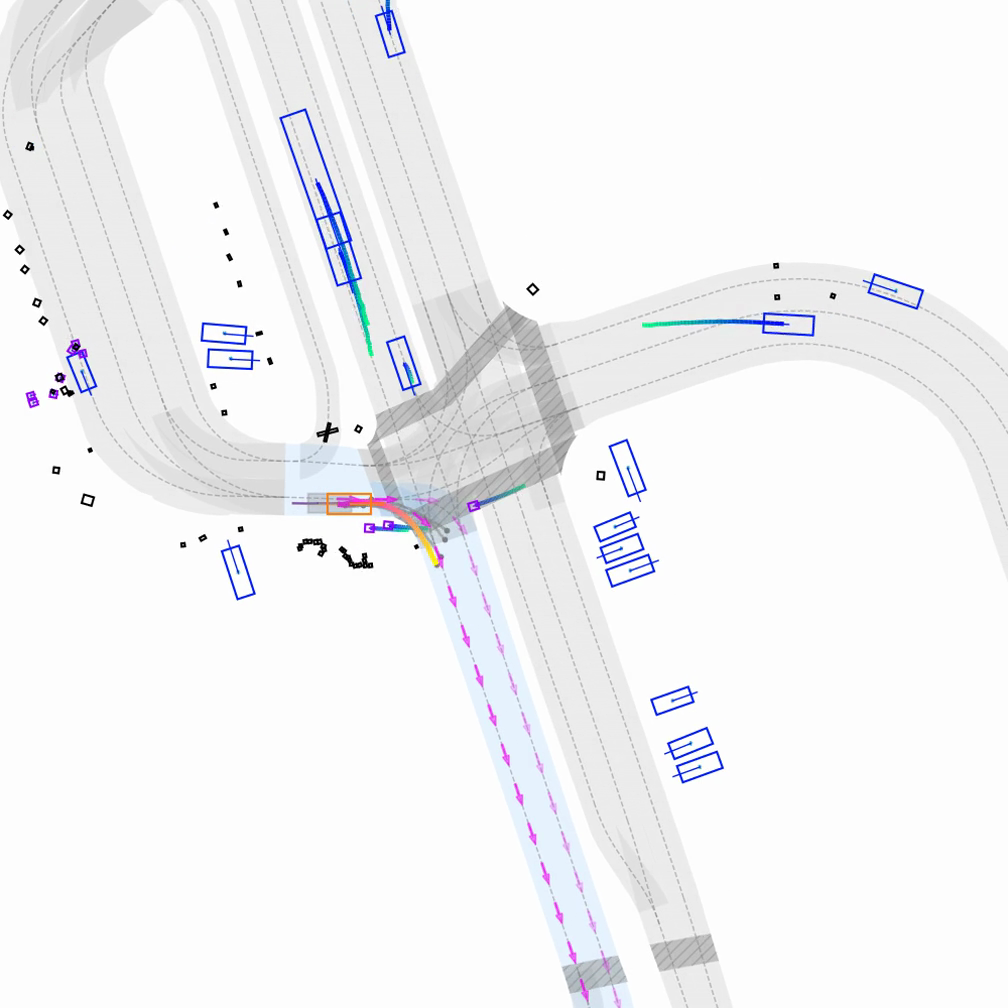}
    \includegraphics[width=0.22\linewidth, trim=300 300 300 300, clip]{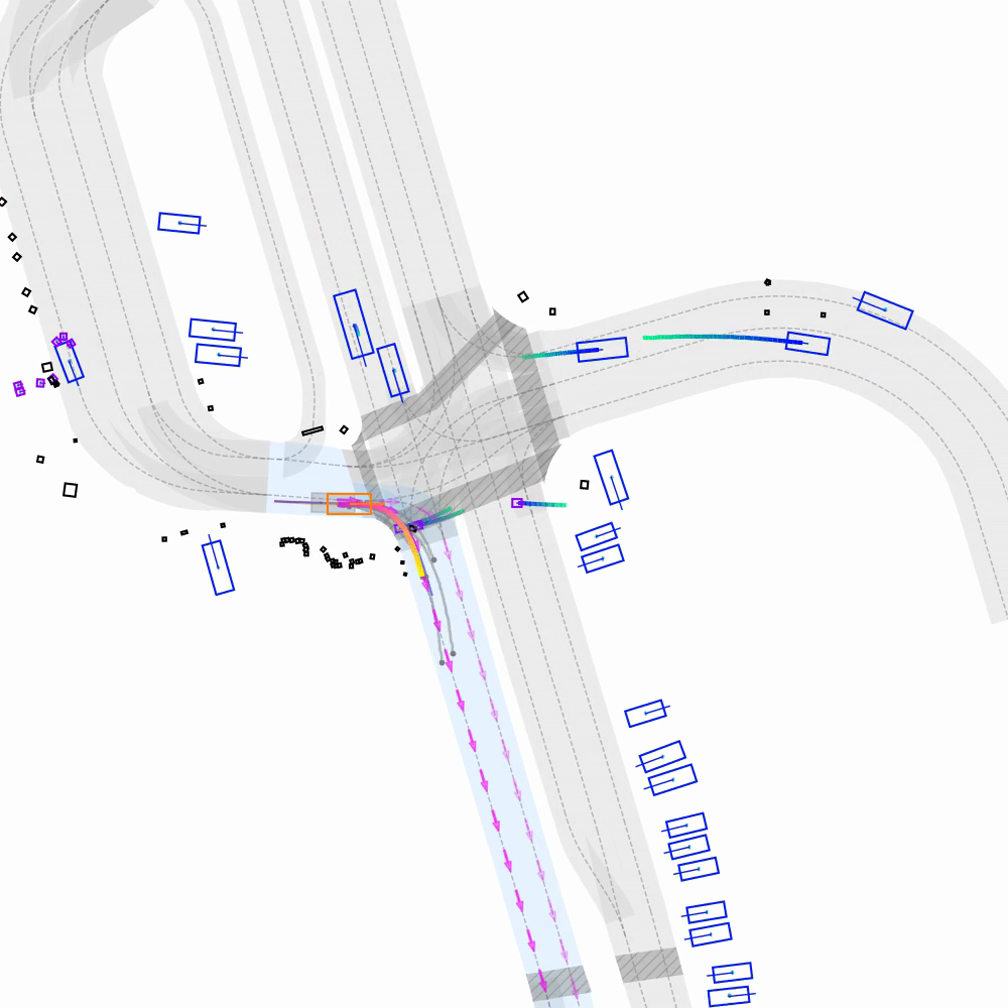}
    \includegraphics[width=0.22\linewidth, trim=300 300 300 300, clip]{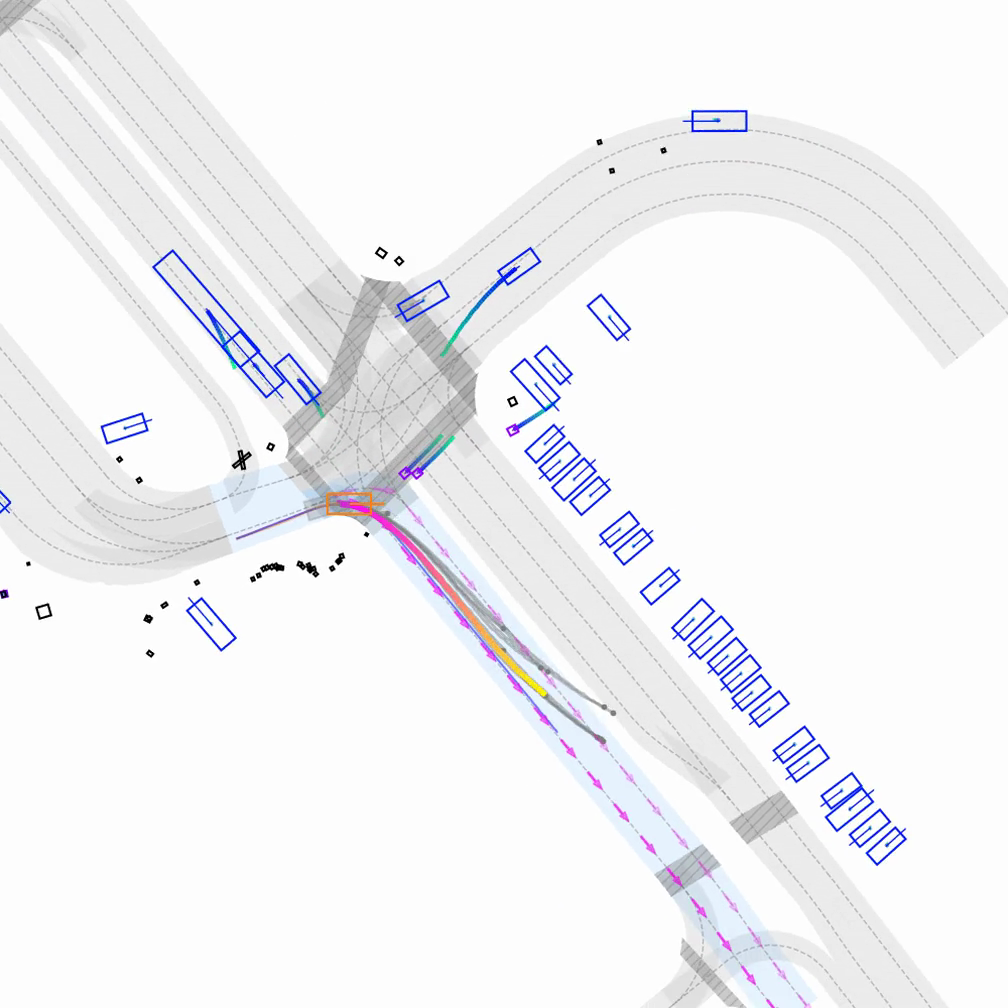}

    \caption{Comparison in a pedestrian-yielding right-turn scenario. The baseline fails to yield to crossing pedestrians and collides, while our method waits for pedestrians to cross and completes the turn safely.}
    \label{fig:simulation_comparison}
    \vspace{-0.5cm}
\end{figure}

This improvement is exemplified in a challenging scenario depicted in Figure~\ref{fig:simulation_comparison}, where a vehicle performs a right turn at an intersection with pedestrians crossing. The baseline model fails to yield to pedestrians and collides with them, demonstrating poor safety awareness. In contrast, our augmented model demonstrates proper pedestrian yielding behavior, waiting for pedestrians to cross safely before completing the turn. Such safety-conscious and interesting behaviors are more prevalent in our augmented dataset due to our sampling and filtering method, which prioritizes vehicles exhibiting dynamic and contextually rich driving patterns. 

\label{sec:experiments}

\begin{figure*}[bt!]
    \centering
    \includegraphics[width=0.9\linewidth]{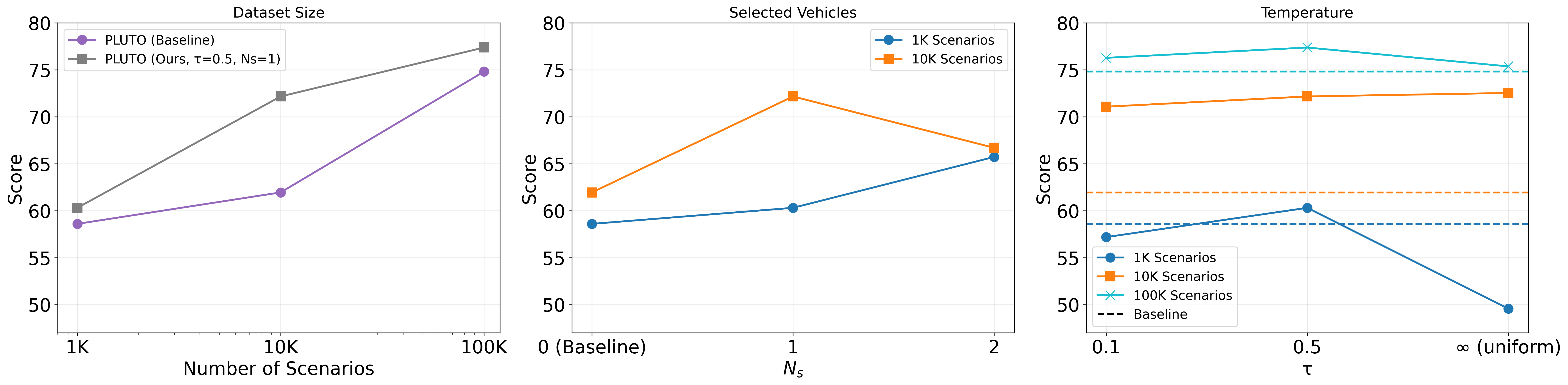}
    \caption{Ablation experiments 1. (1) Dataset size. (2) Number of selected vehicles $N_s$. (3) Temperature parameter $\tau$.}
    \label{fig:ablations}
\end{figure*}

\begin{figure}[b!]
    \centering
    \includegraphics[width=0.9\linewidth]{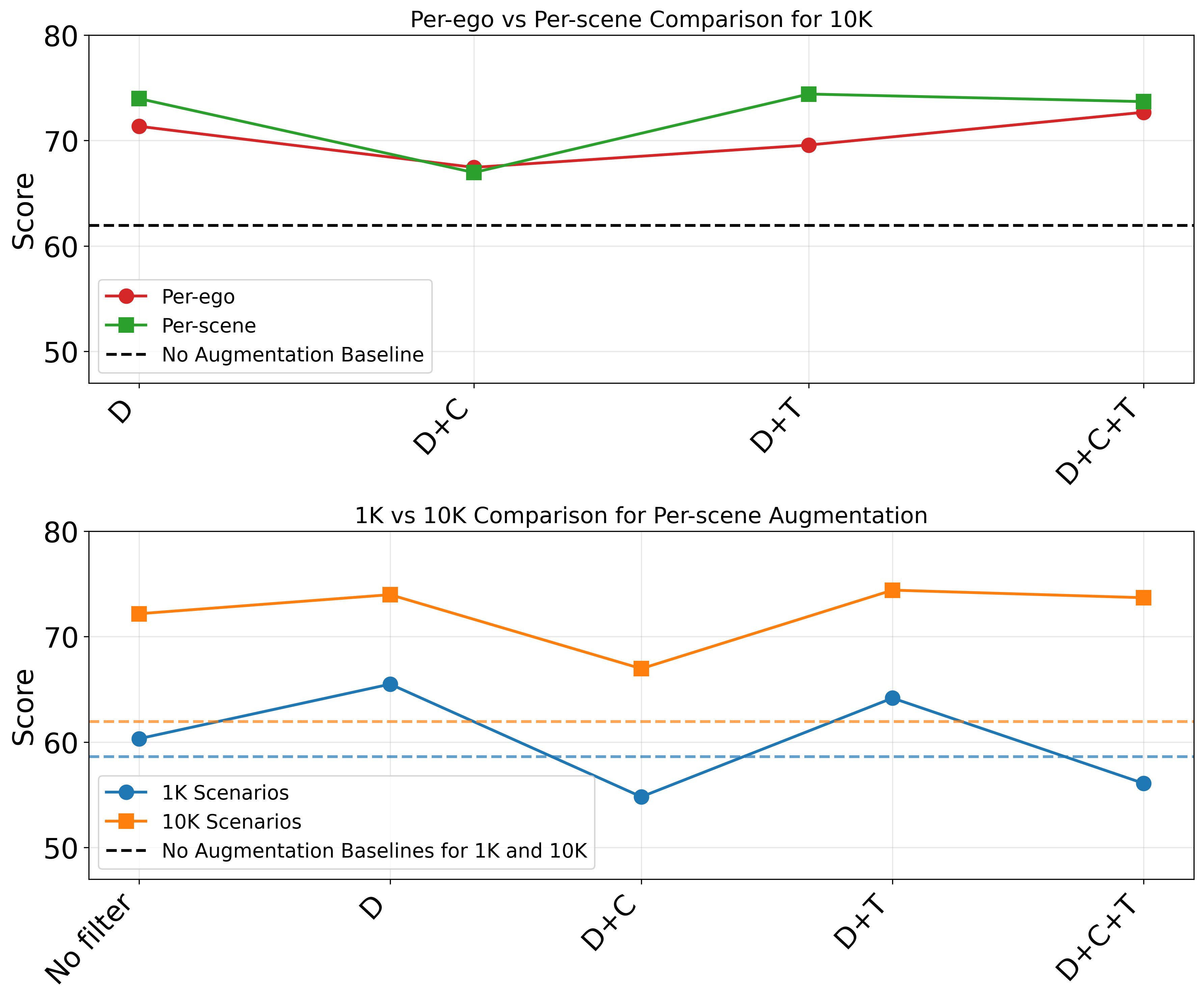}
    \caption{Ablation experiments 2. Top: Per-scene vs per-ego sampling comparison for 10K scenarios. Bottom: Different filter combinations (D=displacement, T=TTC, C=comfort) on 1K and 10K scenarios using per-scene sampling.}
    \label{fig:ablations2}
    \vspace{-0.5cm}
\end{figure}

\subsection{Ablation Studies}

\textbf{Effect of Dataset Size:}  
We conducted ablation studies on dataset size as shown in Figure \ref{fig:ablations}. The results indicate that increasing the dataset size improves performance for both the baseline and our proposed method. Notably, the data augmentation method consistently enhances performance across all dataset sizes, with the most significant gains observed at 10K scenarios. This trend aligns with the expectation that larger datasets generally lead to better generalization and improved model robustness.  

\textbf{Impact of the Number of Selected Vehicles $N_s$:}  
To analyze the effect of augmenting with multiple vehicles, we experimented with selecting 1 or 2 additional vehicles per AV scenario and compared the results against the baseline (no augmentation). In the 1K scenario setting, performance improved when augmenting with 2 vehicles, whereas in the 10K scenario dataset, the best performance was achieved with $N_s=1$. This suggests that while data augmentation is beneficial, excessive augmentation beyond a certain point does not provide further improvements. The results indicate that in low-data regimes, increasing the amount of augmented data is advantageous, but as dataset size grows, additional augmentation yields diminishing improvements.  

\textbf{Influence of the Temperature Parameter $\tau$:} The temperature parameter $\tau$ in the softmax distribution controls sampling bias toward vehicles with higher heading deviations. Lower $\tau$ focuses on dynamic vehicles, while higher $\tau$ leads to more uniform sampling. We tested $\tau=0.1$, $\tau=0.5$, and uniform sampling (\(\tau \to \infty\)). In the 1K dataset, $\tau=0.5$ outperformed others and even exceeded the 10K baseline only using $10\%$ of it, highlighting the benefit of selective augmentation with limited data. However, uniform sampling degraded performance. For 10K and 100K datasets, all $\tau$ values performed similarly, though higher $\tau$ was slightly better for 10K and $\tau=0.5$ for 100K. This suggests $\tau$ selection is crucial in low-data regimes.

\textbf{Per-scene vs Per-ego Sampling:} We compared two sampling strategies for vehicle selection across 10K scenarios, as shown in Figure \ref{fig:ablations2}. Per-scene sampling outperforms per-ego sampling, which is surprising given that per-ego sampling can theoretically select the best vehicles from the entire dataset. However, per-scene sampling introduces more diverse scenarios to the dataset, which could be the primary driver of the observed performance improvement.

\textbf{Effect of Filtering Strategies:} We evaluated different filter combinations (displacement, TTC, and comfort) on both 1K and 10K scenarios using per-scene sampling, as shown in Figure \ref{fig:ablations2}. For these ablations we use $d_{\min}=3$ m, $\kappa_{\text{comf}}=5$, and $\kappa_{\text{ttc}}=0$ violations. Displacement filters consistently improve performance in both dataset sizes, while TTC filters provide additional benefits, particularly in the 10K setting, where displacement + TTC filters achieve performance comparable to the 100K baseline. In the 1K setting, displacement alone provides the best performance, surpassing the 10K baseline while using only 10\% of that dataset. Interestingly, comfort filters generally reduce performance, which might be related to reduced diversity or suboptimal learning of the safety-comfort tradeoff. 

When examining individual metrics in Table \ref{tab:planner_comparison}, we observe that displacement filters improve Progress scores and TTC filters enhance TTC metric performance as expected. These results demonstrate that our filtering approach effectively targets specific driving behaviors and safety metrics.

\section{Discussion}
\label{sec:discussion}

Our approach shows clear performance improvements for imitation learning in autonomous driving with small datasets by selecting suitable trajectories from the surrounding traffic as additional data. We observe that the heading deviation per-scene sampling and displacement and TTC filtering improve performance, likely due to the increased diversity and informativeness of the training trajectories. While our selection criteria are effective and easy to compute for the nuPlan dataset, future work should investigate if more nuanced metrics for selecting trajectories,  such as temporal logic-based filtering~\cite{karagulle2022classification}, can additionally improve the performance.
Additionally, although we demonstrated the effectiveness of our approach on the highly diverse nuPlan dataset, further investigation is needed across other datasets and planning algorithms to assess the generalizability of the observed performance gains. Applying our method to perception-based datasets would also require an additional preprocessing step to generate object-level representations~\cite{Philion2020, li2025bevformer}, which may introduce new challenges related to sensor noise and partial observability.
Lastly, the benefits of our method might vary depending on the baseline model architecture and the specific augmentation parameters, such as the vehicle sampling strategy and softmax temperature. Exploring adaptive augmentation policies and dataset curation strategies could further improve robustness and scalability in future work.

\section{Conclusion}
\label{sec:conclusion}

We propose a data augmentation and filtering strategy using surrounding traffic participants for imitation learning of object-based path planning for autonomous driving. 
Our approach introduces vehicle selection criteria that are efficient to compute and align with the expert trajectory distribution.
We evaluated our method using the PLUTO planner and the nuPlan dataset, demonstrating that the augmentation strategy consistently improves performance for small to medium-sized expert data sets. Our ablations show that using heading deviation as a sampling strategy and displacement and TTC as selection criteria leads to safe driving behaviors, enriching the imitation learning data set.
Beyond autonomous driving, the underlying principle of leveraging observed agent interactions for improved decision-making could extend to other robotics tasks and less structured multi-agent environments, such as aerial traffic or maritime navigation, where large-scale expert demonstrations are difficult to obtain. 

\bibliographystyle{IEEEtran}
\bibliography{root}  

\end{document}